\documentclass[letterpaper, 10 pt, conference]{ieeeconf}

\IEEEoverridecommandlockouts                              
\pdfoutput=1
\usepackage{epsfig}           
\usepackage{mathrsfs}
\usepackage{enumerate}
\usepackage{latexsym}
\usepackage{eufrak}
\usepackage{multirow}
\usepackage{makecell}
\usepackage{array}
\usepackage{graphicx}
\usepackage{epstopdf}

\usepackage{setspace}
\usepackage{url}
\usepackage[linesnumbered,ruled,vlined]{algorithm2e}

\usepackage{color}
\usepackage{amsmath}
\usepackage{amsfonts}
\usepackage{amssymb}
\usepackage{bm}
\usepackage{wrapfig}

\newlength\mylen

\renewcommand \paragraph[1] {\vspace{0.05cm} \textbf{#1}}

\def\etal{\emph{et al.}\xspace}

\newcommand{\argmin}{\operatornamewithlimits{argmin}}

\usepackage{color}

\title{\textbf{Planning Trajectories for Manipulation of Deformable Objects}}
\title{\textbf{Folding Deformable Objects via Trajectory Optimization}}
\title{\textbf{Folding Deformable Objects using \\ Predictive Simulation and Trajectory Optimization}}

\small {
\author{Yinxiao Li, Yonghao Yue, Danfei Xu, Eitan Grinspun, Peter K. Allen
\thanks{All the authors are with Department Computer Science, Columbia University, New York, NY, USA \tt\small \{yli@cs., yonghao@cs., dx2143, eitan@cs. allen@cs.\}columbia.edu}
}
}

\begin{document}
\maketitle
\thispagestyle{empty}
\pagestyle{empty}


\begin{abstract}

Robotic manipulation of deformable objects remains a challenging task.
One such task is folding a garment autonomously.
Given start and end folding positions, what is an optimal trajectory to move the robotic arm to fold a garment?
Certain trajectories will cause the garment to move, creating wrinkles, and gaps, other trajectories will fail altogether.
We present a novel solution to find an optimal trajectory that avoids such problematic scenarios.
The trajectory is optimized by minimizing a quadratic objective function in an off-line simulator, which includes material properties of the garment and frictional force on the table. 
The function measures the dissimilarity between a user folded shape and the folded garment in simulation, which is then used as an error measurement to create an optimal trajectory.
We demonstrate that our two-arm robot can follow the optimized trajectories,
achieving accurate and efficient manipulations of deformable objects.

\end{abstract}

\section{Introduction}
\label{sec:intro}

\begin{figure}[t]
\begin{center}
  \centering
  \includegraphics[width=0.48\textwidth]{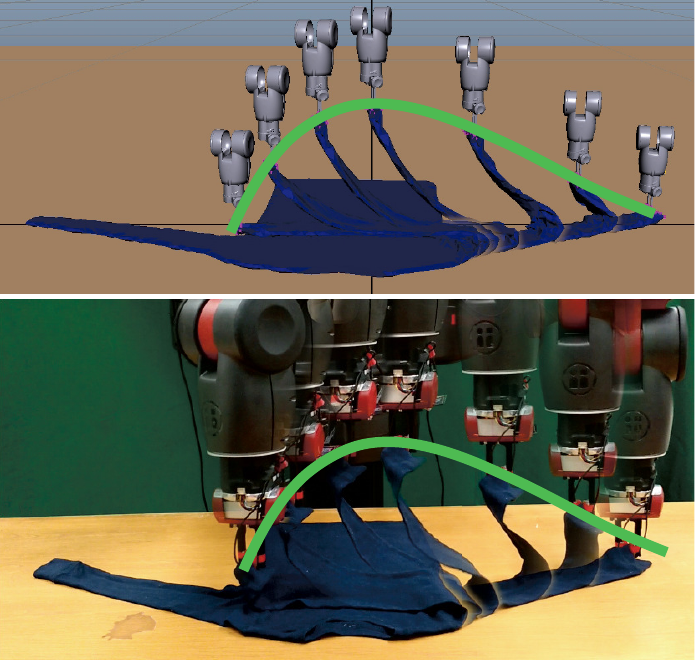}
\end{center}
   \caption{
	Comparison of our simulation of robotic manipulation ({\sc{TOP}}) and real
robot implementation ({\sc{BOTTOM}}). The green curves show the virtual and the real trajectories for folding.}
\label{fig:intro}
\end{figure}

Robotic folding of a garment is a difficult task because it requires sequential manipulations of a highly unconstrained, deformable object.
Given the garment shape, the robot can fold it by following a folding plan~\cite{millerICRA2011}\cite{Milleretal_IJRR2012}.
However, the layout of the same folding action can vary in terms of the material properties such as cloth hardness and the environment such as friction between the garment and the table.
Given the starting and ending folding positions, different folding trajectories will lead to different results.
In this paper, we propose a novel method that learns optimal folding trajectory parameters from predicted thin shell simulations of similar garments, which can then be applied to a real garment folding task (see Figure~\ref{fig:intro}). The contributions of our paper are: 

\begin{itemize}
\item[-] A fast and robust algorithm that can detect garment key points such as sleeve ends, collar, and waist corner, automatically. These key points can be used for folding plan generation.
\item[-] An online optimization algorithm that learns optimal trajectories for manipulation from mathematical model evolution combined with predictive thin shell simulation.
\item[-] A novel approach that adjusts the simulation environment to the robot working environment for the purpose of creating a similar manipulation result. 
\item[-] The trajectories are general in that they can be scaled to accommodate similar garments of different size.
\item[-] Experimental results with a Baxter robot showing successful folding trajectories for a number of different garments including sweaters, pants, and towels.
\end{itemize}

\begin{figure*}[!htb]
  \centering
  \includegraphics[width=0.99\textwidth]{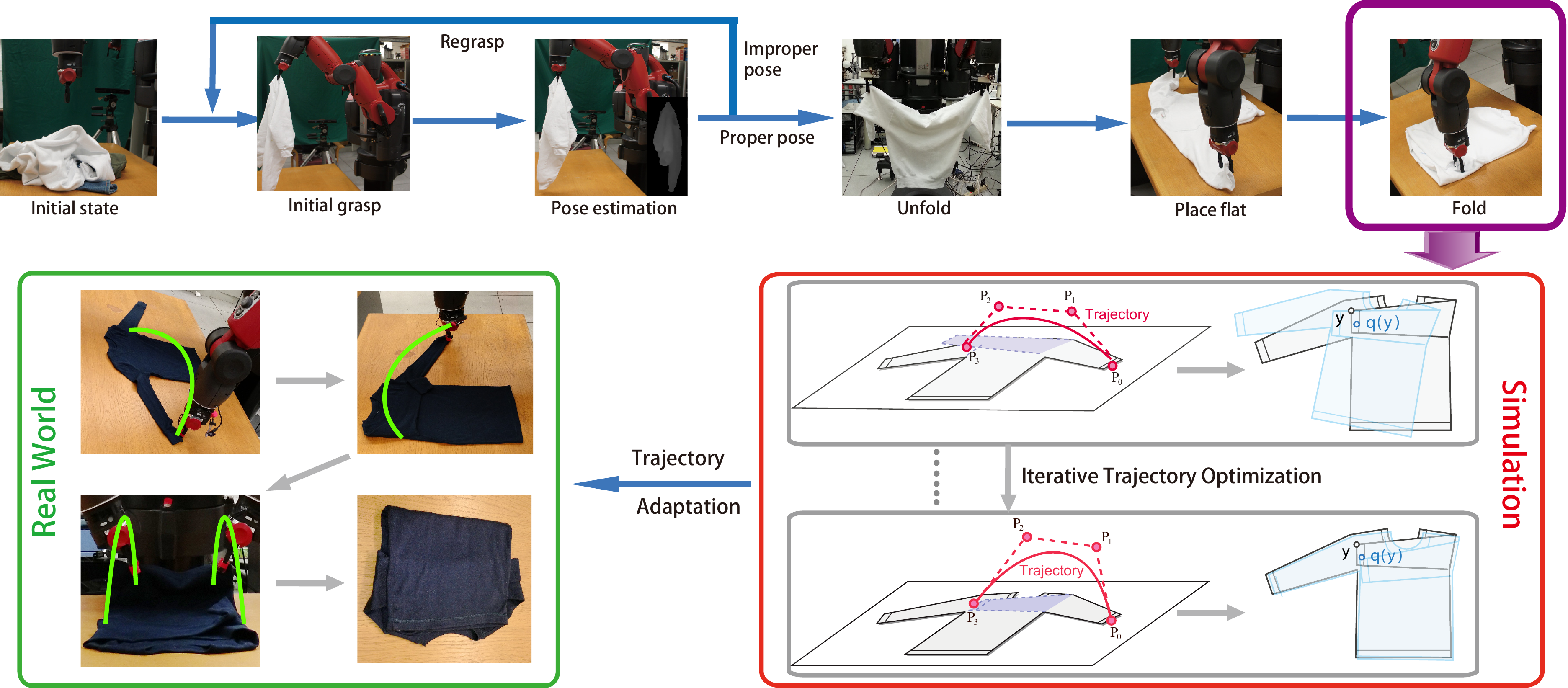}
  \caption{
    {\sc Top Row: } The entire pipeline of dexterous manipulation of deformable objects.
    In this paper, we are focusing on the phases of garment folding, as highlighted in the purple rectangle.
    {\sc Bottom Row: }  Details of the folding procedure. We apply off-line simulation with iterative trajectory optimization to find the best trajectory for a specific folding action by comparing the result (light blue contour) with template (black contour). Similar steps are repeated until the garment is folded in the simulator. Then all the folding trajectories are exported, adapted, and implemented on a real robot. Green arcs illustrate the actual trajectories of robotic arms.}
\label{fig:flowchart}
\end{figure*}

Figure~\ref{fig:flowchart} shows the complete pipeline of garment manipulation, as well as the key steps of garment folding.
The garment folding is the final step of the entire pipeline of garment manipulation which contains grasping, visual recognition, regrasping, unfolding, placing flat, and folding.
Our previous work~\cite{LiICRA2014}\cite{LiIROS2014}\cite{LiICRA2015} has successfully addressed all the stages of the pipeline with the exception of the final folding task.
This paper specifically addresses the robotic folding task (purple rectangle in Figure~\ref{fig:flowchart}) with the goal of finding optimal trajectories to successfully fold garments.

We begin by assuming the garment is placed flat on the table initially, as shown in our previous work~\cite{LiICRA2015}.
By detecting the key points of the garment (see section~\ref{sec:localization}), a pre-defined folding plan is used to create optimal trajectories for folding the garment.
After several steps, we obtain a desired folding result in the real world using the Baxter robot, which is comparable to the result from the simulation.

\section{Related Work}
\label{sec:relatedwork}

There are many challenges associated with the manipulation of a deformable object such as a garment.
One of the challenges is unfolding a garment from an arbitrary state and placing it flat on a table.
Many researchers start with simple garments such as a towel~\cite{Shepard2010}.
By iteratively looking for the lowest corner point of the towel, the robot is able to unfold it and place flat on a table. 
Then the towel can be quickly folded by symmetric information.
For more complicated garments such as sweaters and pants, their states (poses) have to be recognized first by either image-based perception~\cite{Willimon2013}\cite{LiICRA2014}\cite{DoumanoglouECCV2014} or 3D shape matching~\cite{Kita2009}\cite{LiIROS2014}, etc..
Deformable objects such as a garment have large dimensional state spaces which are hard to track and recognize. 
Therefore, for such tasks, many researchers employ a large database which contains exemplars of different states of an object from off-line simulation or real garments as the training data. 
By using SIFT feature~\cite{Willimon2013}\cite{LiICRA2014} or volumetric features~\cite{DoumanoglouECCV2014}\cite{LiIROS2014}, the state of a garment is recognized and tracked. 
After regrasping several times, the garment can be unfolded and placed flat on a table by a dual-arm robot~\cite{Towner2011}\cite{StriaIROS2014}\cite{LiICRA2015}, a prerequisite for garment folding.

With the garment fully spread on the table, attention is turned to parsing its shape.
S. Miller~\etal have designed a parametrized shape model for unknown garments~\cite{millerICRA2011}\cite{Milleretal_IJRR2012}.
Each set of parameters defines a certain type of garment such as a sweater or a towel.
The contour-based garment shape model was further improved by J. Stria~\etal using polygonal models~\cite{Stria2014TAROS}.
The detected garment contour is matched to a polygonal model by removing non-convex points using a dynamic programming approach.
Folding is the ultimate goals of garment manipulation and only a few researchers have achieved this.
F. Osawa \etal used a robot to fold a garment with a special purpose table that contains a plate that can bend and fold the clothes assisted by a dual-arm robot.
Another folding method using a PR2 robot was implemented by J. van den Berg \etal~\cite{vandenberg2010}.
The core of their approach was the geometry reasoning with respect to the cloth model without any physical simulation.
Similar work done by J. Stria~\etal~\cite{StriaIROS2014} using two industrial arms and a polygonal contour model.
Kita,~\etal~\cite{Kita2014} use a humanoid robot to fold a garment starting from a random configuration.

None of the previous works focus on trajectory optimization for garment folding, which brings uncertainty to the layout given the same folding plan.
One possible case is that the garment shifts on the table during one folding action so that the targeted folding position is also moved.
Another case is that an improper folding trajectory causes additional deformation of the garment itself, which can accumulate.

\begin{figure}[!htpb]
\begin{center}
 \includegraphics[width=0.49\textwidth]{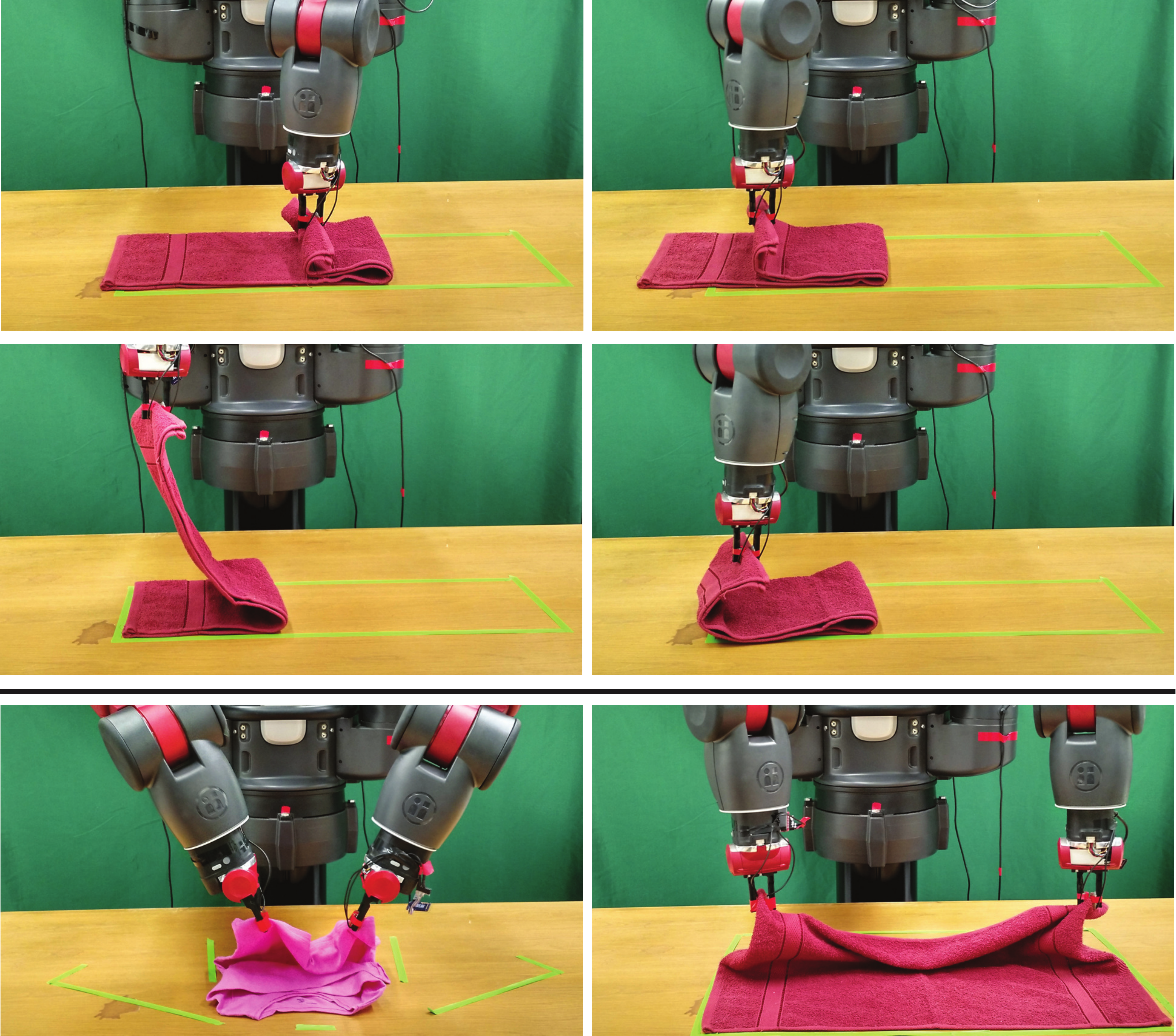}
\end{center}
   \caption{Failure example with improper folding trajectories.
	{\sc{First Row:}} Folding trajectory is low and flat that causes drift to the towel and long-sleeve T-Shirt.
	{\sc{Second Row:}} Folding trajectory is too high when the gripper approaching the target folding position that piles up the towel.
	{\sc{Third Row:}} Dual-arm folding. If the distance between the two arms is too close, the folding may fail.}
\label{fig:failure_examples}
\end{figure}

Figure~\ref{fig:failure_examples} shows a few failure examples with improper trajectories.
We use green tape on the table to show the original position of the garments.
The first two rows show that if the moving trajectory is too low and close to the garment, the folded part will fall down, pull the rest, and cause drift of the whole garment. 
These cases usually happen when the folding step is lengthy without trajectory optimization.
The third row shows a case where the folding trajectory is too high, which will cause extra wrinkles or even piling up.
The last row shows two cases using two arms to fold.
If the arms are close, the part in between loses tension, and will fall down and pull the rest away.
The focus of this paper is to create trajectories for folding that will overcome these problems.

\section{Simulation Environment}
\label{section:Simulation}

\subsection{Folding Pipeline in Simulation}
In the model simulation, we use a physics engine~\cite{urlMaya} to simulate the movement and deformation of the garment mesh models.
We assume there is only one garment for each folding task, which has been placed flat on a table.
A virtual table is added to the scene which the garment lies on, as shown in Figure~\ref{fig:intro}, top.

During each folding step, the robot arm picks up a small part of the mesh, moves it to the target position following a computed trajectory, and places it on the table to simulate an entire folding scenario.
If the part of the garment to be folded is relatively wide, then both left and right arms may be involved.
The trajectory is generated using a B\'ezier curve, which will be discussed in section~\ref{sec:trajectory_optimization}.

Most of the garment mesh models are built from our test garments. 
A garment mesh is created by first extracting the contour of the garment (see section~\ref{sec:localization}).
Then by inserting points on the inside of the garment contour, we triangulate a mesh by connecting these points.
Lastly, we mirror the mesh to construct a two-sided garment mesh.


\subsection{Parameter Adaptation}

There are two key parameters needed to accurately simulate the real world folding environment. 
The first is the material properties of the fabric, and the second is the frictional forces between the garment and the table.

\subsubsection{Material properties}
Through many experiments, we found that the most important property for the garments in the simulation environment is shear resistance.
It specifies the amount the simulated mesh model resists shear under strain; when the garment is picked up and hung by gravity, the total length will be elongated due to the balance between gravity force and shear resistance.
An appropriate shear resistance measure allows the simulated mesh to reproduce the same elongation as the real garment.
This parameter will bridge the gap between the simulation and the real world for the garment mesh model.

\begin{figure}[!htpb]
\begin{center}
 \includegraphics[width=0.48\textwidth]{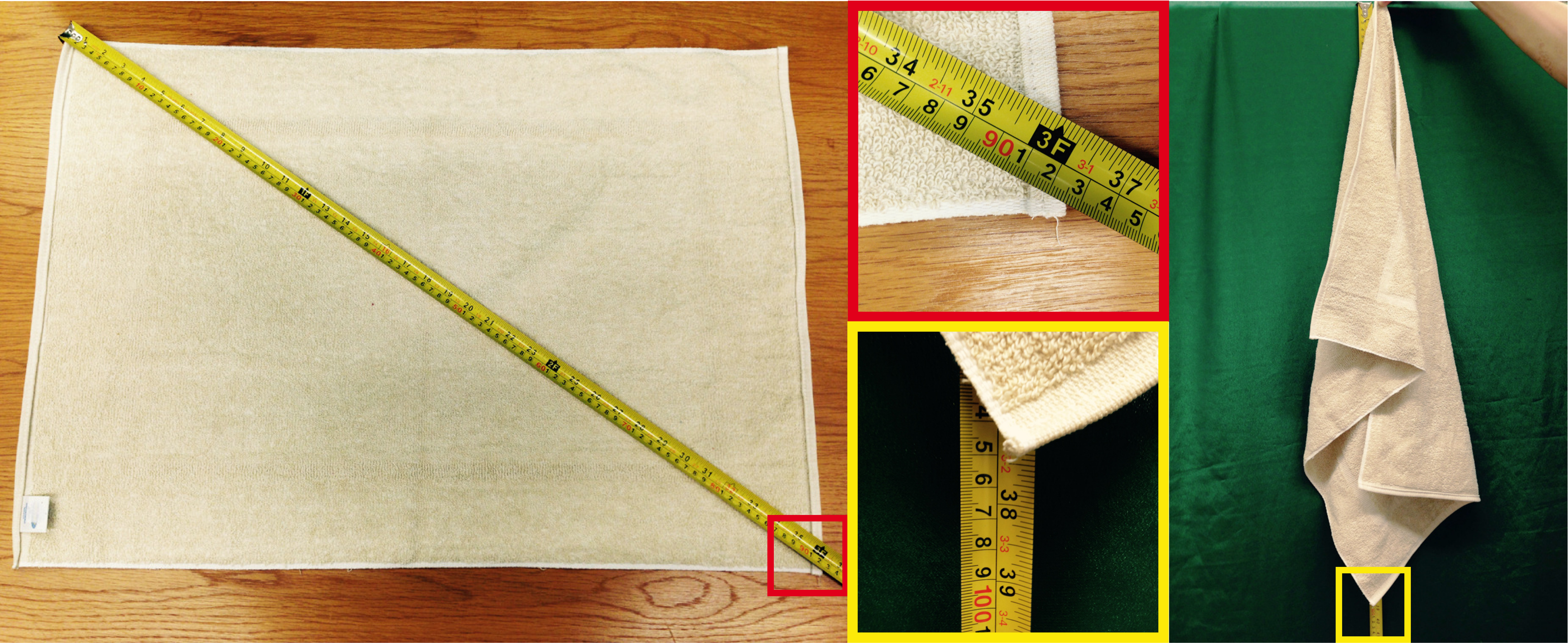}
\end{center}
   \caption{Method for measuring the shear resistance.
	{\sc{Left:}} Diagonal length measurement.
	{\sc{Middle:}} Zoomed in regions.
	{\sc{Right:}} The garment is hanging under gravity.}
\label{fig:measurement}
\end{figure}

For each garment, we follow the steps described below to measure the shear resistance. 
Figure~\ref{fig:measurement} shows an example.
\begin{itemize}
\item[-] Manually pick one extremum part of the garment such as the sleeve end of a T-shirt, the waist part of a pair of pants, and a corner of a towel. 
\item[-] Hang the garment under gravity and measure the length between the picking point and the lowest point as $L_1$
\item[-] Slowly put down the garment on a table and keep the picking point and the lowest point in the previous step at maximum spread condition. Measure the distance between these two points again as $L_2$. The shear resistance fraction is defined as the following
\begin{align}
shear\_frac = ({L_1} - {L_2})/{L_2}
\end{align}
\item[-] We then pick up and hang the virtual garment in \emph{Maya}, adjusting the \emph{Maya} shear parameter such that the shear fraction as calculated in the simulator is identical to the real world. 
\end{itemize}

\subsubsection{Frictional forces}
The surface of the table can be rough if covered by a cloth sheet or slippery if not covered, which leads to variance in friction between the table and garment.
A shift of the garment during the folding can possibly impair the whole process and cause additional repositioning.
Adjusting the frictional level in the simulation to the real world is crucial and necessary for trajectory optimization.

To measure the friction between the table and the garment, we do the following steps.
\begin{itemize}

\item[-] Place a real garment on the real table of length $L_t$.
\item[-] Slowly lift up one side of the real table, until the garment in the real world begins to slide. The lifted height is $H_s$. The friction angle is computed as,
\begin{align}
{\angle _{Friction}} = {\sin ^{ - 1}}({H_s}/{L_t})
\end{align}
\item[-] In the virtual environment, the garment is placed flat on a table with gravity. Assign a relatively high friction value to the virtual table. Lift up one side of the virtual table to the angle of ${\angle _{Friction}}$.
\item[-] Gradually decrease the frictional force in the virtual environment until the garment begins to slide. 
Use this frictional force in the virtual environment as it mirrors the real world

\end{itemize}

With these two parameters, we obtain similar manipulation results for both the simulation and the real garment.

\section{Trajectory Optimization}
\label{sec:trajectory_optimization}

The goal of the folding task is specified by the initial and folded
shapes of the garment, and by the starting and target positions of the
grasp point (as in Figure~\ref{fig:traj_opt}). Given the simulation 
parameters, we seek the trajectory that effects the desired set of folds. 
We first describe how to optimize the trajectory for a single end 
effector, and then discuss the case of two end effectors.

\subsection{Trajectory parametrization}
We use a B\'ezier curve~\cite{Farin:1988} to describe the trajectory.
An $n$-th order B\'ezier curve $\mathbf{T}(u)$ has $(n+1)$ control points 
$\mathbf{P}_k = ({P}_{k,x}, {P}_{k,y}, {P}_{k,z})^T\in \mathbb{R}^3$, defined by
\begin{align}
\mathbf{T}(u) = \sum_{k=0}^n B_k^n(u) \mathbf{P}_k,
\end{align}
where $B_k^n(u) = \begin{pmatrix} n \\ k \end{pmatrix} (1-u)^{n-k} u^k$ are the Bernstein basis
functions.

\begin{figure}[!htpb]
\centering
\includegraphics[width=0.98\linewidth]{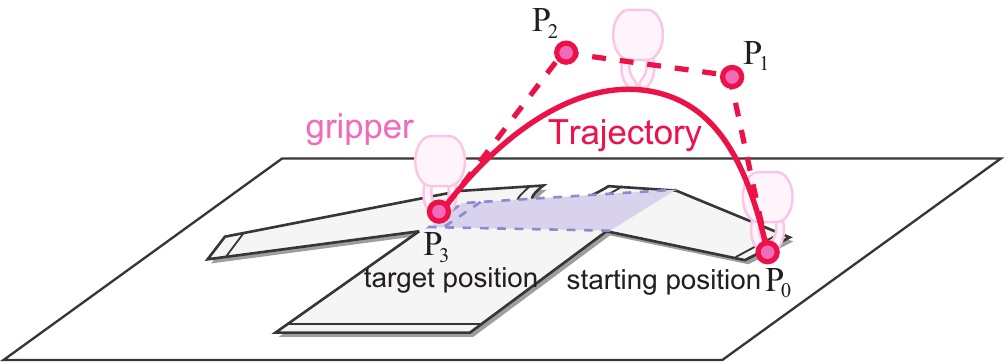}
\caption{An example of the folding task: we want to fold a sleeve into the blue
target position, by using a robotic gripper to move the tip of the sleeve 
(grasp point) from the starting position ($\mathbf{P}_0$) to the target position
($\mathbf{P}_3$), following a trajectory, shown as the red curve.
$\mathbf{P}_1$ and $\mathbf{P}_2$ are knot points that form the B\'ezier trapezoid.}
\label{fig:traj_opt}
\end{figure}

We use $n=3$ for simplicity, but our method can be easily extended to
deal with higher order curves. $\mathbf{P}_0$ and $\mathbf{P}_3$ are
fixed to the specified starting and target positions of the grasp
point (as in Figure~\ref{fig:traj_opt}). The intermediate control points $\mathbf{x} =
(\mathbf{P}_1^T, \mathbf{P}_2^T)^T$ can then be adjusted to define a new trajectory using the objective function defined below. The update rule is described in section~\ref{sec:optimization}.
\begin{align}
\label{eq:objective}
\mathbf{x}_{opt} = \argmin_{\mathbf{x}} \{\underbrace{ l_\mathbf{x} + \alpha D(\mathcal{S}_t, \mathcal{S}_{\mathbf{x}}) }_{C(\mathbf{x})} \}^2.
\end{align}

\begin{wrapfigure}{r}{0.2\textwidth}
  \begin{center}
\includegraphics[width=1.0\linewidth]{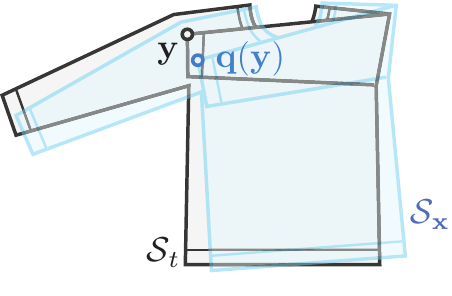}
  \end{center}
  \caption{The dissimilarity captures the misalignment between 
$\mathcal{S}_t$ and $\mathcal{S}_{\mathbf{x}}$ by integrating the distance 
between the corresponding points $\mathbf{y} \in \mathcal{S}_t$ and 
$\mathbf{q}(\mathbf{y}) \in \mathcal{S}_{\mathbf{x}}$ over the garment.}
\label{fig:dissimilarity}
\end{wrapfigure}

Here $C(\mathbf{x})$ is a cost function with two terms. The first term
penalizes the trajectory length 
$l_\mathbf{x}$, thus preferring a folding path that is efficient in
time and energy. The second term seeks the desired fold, by penalizing dissimilarity $D(\mathcal{S}_t, \mathcal{S}_{\mathbf{x}})$
between the desired folded shape $\mathcal{S}_t$, compared to
the shape $\mathcal{S}_{\mathbf{x}}$ obtained by the
candidate folding trajectory $\mathbf{x}$, as predicted by a cloth simulation; we used a physical simulation engine~\cite{urlMaya}, 
for the cloth simulation.
The weight $\alpha$ balances the two terms; we used $\alpha 
= 10^3$ in our experiment. 
See section~\ref{sec:optimization} for optimization details.

Intuitively, dissimilarity measures the difference between the desired folded shape and the folded garment in simulation.
We define the dissimilarity term as
\begin{align}\label{eq:dissimilarity-cont}
D(\mathcal{S}_t, \mathcal{S}_{\mathbf{x}}) = 
\frac{1}{|\mathcal{S}_t|}\int_{\mathcal{S}_t} \|\mathbf{q}(\mathbf{y}) - 
\mathbf{y}\| dA,
\end{align}
where $|\mathcal{S}_t|$ is the total surface area of the garment mesh including both sides of the garment, $\mathbf{y} \in
\mathcal{S}_t$ is a point on the target folded shape $\mathcal{S}_t$,
$\mathbf{q}(\mathbf{y}) \in \mathcal{S}_{\mathbf{x}}$ is the corresponding
point on the simulated folded shape, and $dA$ is the area measure, see 
Figure~\ref{fig:dissimilarity}. 
Our implementation assumes $\mathcal{S}_t$ 
and $\mathcal{S}_{\mathbf{x}}$ are given as triangle meshes, and 
discretizes~\eqref{eq:dissimilarity-cont} as
\begin{align}
\tilde{D}(\mathcal{S}_t, \mathcal{S}_{\mathbf{x}}) = 
\frac{1}{|\mathcal{S}_t|}\sum_i \|\mathbf{q}_i - \mathbf{y}_i\| A_i,
\end{align}
where $\mathbf{y}_i$ is the barycenter of $i$-th triangle on the target shape, 
$\mathbf{q}_i$ is the (corresponding) barycenter of $i$-th triangle on the simulated
shape, and $A_i$ is the area of the $i$-th triangle on the target shape.

To compute the trajectory length $l_\mathbf{x}$, we use the De Casteljau's
algorithm~\cite{Farin:1988} to recursively subdivide the B\'ezier curve 
$\mathbf{T}$ into a set of B\'ezier curves $\mathbf{T}^{(j)}$, until the 
deviation between the chord length ($\|\mathbf{P}_0^{(j)} - 
\mathbf{P}_3^{(j)}\|$) and the total length between the control points 
($\sum_{i=0}^2 \|\mathbf{P}_i^{(j)} - \mathbf{P}_{i+1}^{(j)}\|$) for each 
subdivided curve $\mathbf{T}^{(j)}$ is sufficiently small. Then, $l_\mathbf{x}$
is approximated by summing up the chord lengths of all the subdivided curves: 
$l_\mathbf{x} \approx \sum_j \|\mathbf{P}_0^{(j)} - \mathbf{P}_3^{(j)}\|$.

We initialize $\mathbf{P}_1$ and $\mathbf{P}_2$ as
\begin{align}
\mathbf{P}_1 = \frac{2}{3}\mathbf{P}_0 + \frac{1}{3}\mathbf{P}_3 + h \|\mathbf{P}_0-\mathbf{P}_3\| \mathbf{e}_v,\\
\mathbf{P}_2 = \frac{1}{3}\mathbf{P}_0 + \frac{2}{3}\mathbf{P}_3 + h \|\mathbf{P}_0-\mathbf{P}_3\| \mathbf{e}_v,
\end{align}
where $\mathbf{e}_v$ is the unit vector in the upward vertical direction,
$h$ is a constant value of $1/3$, which means the initial trajectory will have equal horizontal extent between knot points.

\subsection{Optimization.}
\label{sec:optimization}
To optimize equation \eqref{eq:objective}, we apply a secant version of
the Levenberg-Marquardt algorithm~\cite{Madsen:2004}\cite{Nocedal:2006}. 
For the current trajectory generated by $\mathbf{x}$, we estimate the derivative $\nabla C(\mathbf{x})$ of
the cost function $C(\mathbf{x})$ numerically, by sampling slightly modified
trajectories $\mathbf{x} + \delta \mathbf{e}_j$, where $\mathbf{e}_j, 
1\leq j \leq \textrm{dim}(\mathbf{x})$, are the orthonormal bases, and we used 
$\delta = 10^{-1}$ in our implementation.

The secant version of Levenberg-Marquardt algorithm iteratively builds a local 
quadratic approximation of $\{C(\mathbf{x})\}^2$ based on the numerical
derivative, and then takes a step toward an improved state. The direction of
the step is a combination of the steepest gradient descent direction and the 
conjugate gradient direction. We use the specific approach described
by Madsen et al.~\cite{Madsen:2004} (see \S3.5 therein).
The iterative procedure terminates when the improvement in $\{C(\mathbf{x})\}^2$ becomes
sufficiently small.

\subsection{Multiple arms.}
In the case of using multiple arms, we associate an individual trajectory 
$\mathbf{x}_i$ to each of the arms $R_i$. We then extend the state variable
to $\mathbf{x} = (\mathbf{x}_1^T, ...)^T$. The rest of the optimization
procedure is the same as the single arm case. 
Note that both single and dual-arm trajectories are in 3D space.
The optimization for dual-arm trajectories is able to find a solution which will overcome failures such as shown in Figure~\ref{fig:failure_examples} bottom.

\section{Application to Garment Folding}

\subsection{Key Points Localization}
\label{sec:localization}

\begin{figure}[!htpb]
  \center
    \includegraphics[width=0.48\textwidth]{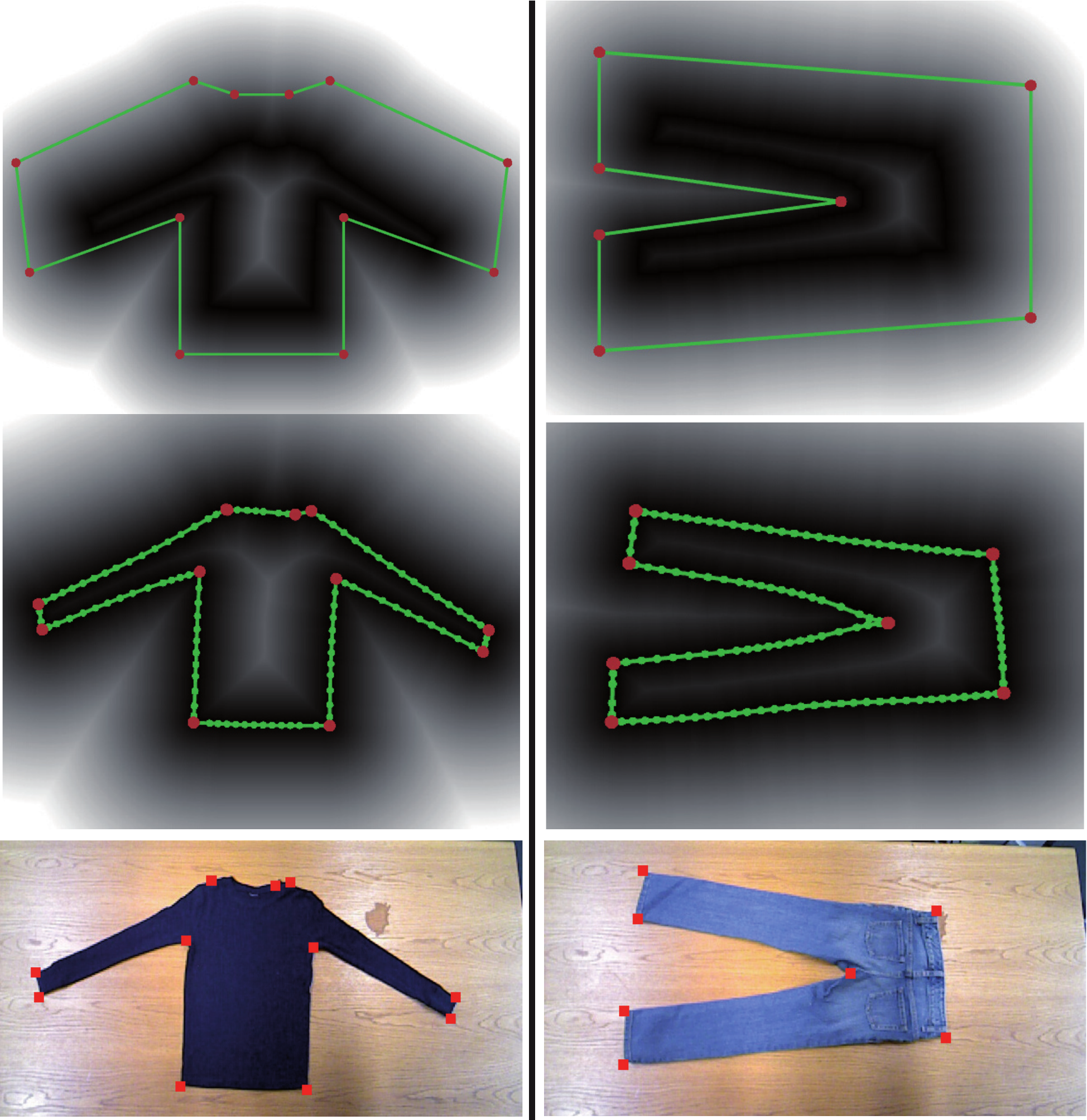}
   \caption{ Red dots are the predefined key points for the garment, such as sleeve ends.
	Left column is a long-sleeve t-shirt example, and the right column is a pants example.
	{\sc{Top:}} Initialized contour. Each garment category is initialized with a different contour.
	{\sc{Middle:}} Fitting results. The contour shrinks onto the boundary of the garment according to the distance field. 
	{\sc{Bottom:}} Fitting results mapped back to the original input image.}
\label{fig:garment_snakes}
\end{figure}

We assume the garment is placed roughly in the center of the table, as shown in the Figure~\ref{fig:garment_snakes}, bottom.
Our first step is to segment the garment from the background.
Since we can easily obtain a table or a surface with homogeneous color~\cite{millerICRA2011}\cite{Stria2014TAROS}, a fast color-based supervised image segmentation method is suitable for our task.
We apply a marker-based watershed method~\cite{vincent1991} with the four corners of the table initialized as background and the center initialized as the foreground.
More complicated scenes can employ advanced image segmentation algorithms such as GrabCut~\cite{Rother_grabcut}.

To register the feature points, such as the corner points of sleeves,
we employ a 2D registration technique to register a pre-defined garment template
(as in Figure~\ref{fig:garment_snakes} top row) with the captured 
garment mask. Our 2D registration is based on our 3D non-rigid 
registration code for thin shell models~\cite{LiICRA2015}, 
and can deal with garment masks that have curved contours.

We first initialize a distance field using the segmented mask from 
the previous step. 
The category of the garment can be easily recognized by using a template matching algorithm, which leads to an associated garment initial template. 
We register the template with the garment mask by minimizing the following energy function:
\begin{align}
\label{eqn:energy}
E_{T}(S,\bar{S}) = E_{\textrm{fit}}(S,T) + \kappa E_{\textrm{stretch}}(\bar{S}, \mathbf{x}) 
+ \beta E_{\textrm{bend}}(\bar{S}, \mathbf{x}),
\end{align}
where $T$ is the target garment mask, $S$ is the current deformed 2D
contour, and $\bar{S}$ is a reference 2D contour, which is from the previous state. 
$E_{\textrm{fit}}(S,T)$ penalizes discrepancies between the
contour and the target mask.
The last two terms seek 
to limit and regularize the deformation of the contour in order to 
preserve the angle features,
which includes the stretching and bending energies, weighted by
user specified coefficients $\kappa$ and $\beta$.

We represent the contour as a closed loop consisting of line segments.
In this discrete representation, $E_{\textrm{fit}}$ is computed as
\begin{align}
\tilde{E}_{\textrm{fit}}(S,T) = \sum_i \left( \textrm{dist}_T(\mathbf{x}_i) \right)^2 l_i,
\end{align}
where $l_i$ and $\mathbf{x}_i$ are the length and the midpoint of the $i$-th 
line segment of the deformed contour.
$E_{\textrm{stretch}}$ is computed as
\begin{align}
\tilde{E}_{\textrm{stretch}}(\bar{S}, \mathbf{x}) = \frac{1}{2} \sum_i \left( \frac{l_i}{\bar{l}_i} - 1\right)^2 \bar{l}_i,
\end{align}
where $\bar{l}_i$ is length of the $i$-th line segment of the reference contour.
$E_{\textrm{bend}}$ is computed as
\begin{align}
\tilde{E}_{\textrm{bend}}(\bar{S}, \mathbf{x}) = \frac{1}{2} \sum_j \left( \frac{\theta_j}{\bar{\theta}_j} -1 \right)^2 \bar{L}_j,
\end{align}
where $\theta_j$ and $\bar{\theta}_j$ are the angles between the adjacent 
line segments at $j$-th vertex of the deformed and the reference contours, 
respectively. $\bar{L}_j$ are the average length of the adjacent line segments
at $j$-th vertex of the reference contour.

Our registration iteratively updates $S$ by using the secant version of
the Levenberg-Marquardt algorithm~\cite{Madsen:2004}\cite{Nocedal:2006}.
Initially, each vertex in the template polygon is marked as a \textit{
feature point} (red points in Figure~\ref{fig:garment_snakes}, top row), and is
assigned a unique ID to identify its semantic meaning. 
At the end of each step, we subdivide each contour segment if its length is
larger than a threshold, and mark the newly added vertex as 
\textit{non-feature points} (green points in 
Figure~\ref{fig:garment_snakes}, middle row). 
Next, we merge any pair of adjacent segments if they do not share a feature point and their length is below a threshold. 
The subdivide and merge operations guarantee that the garment 
contour is sufficiently but not overly sampled. Then, we update the
reference contour $\bar{S}$ by $S$. 

We repeat the iteration until
the reduction in $E_{T}(S,\bar{S})$ becomes sufficiently small.
Finally, the positions as well as the semantic meanings of the feature
points in the garment mask are identified by retrieving the feature points via the unique point ID
in the registered contour.


\section{Experimental Results}
\label{section:experiments}

To evaluate our results, we tested our method on several different garments such as long-sleeve t-shirts, pants, and towels for multiple trials, as shown in Figure~\ref{fig:garment_stat} left.
These garments require both single and dual-arm folds.
A high resolution video of our experimental results is online at {\url{http://www.cs.columbia.edu/~yli/IROS2015}}.

\subsection{Robot Setup}
In our experiments, we use a Baxter research robot, which is equipped with two arms with seven degrees of freedom. 
We mount a Prime Sense Xtion range sensor~\cite{primesense} on top of the Baxter head panel, which has been calibrated to the robot base frame.
To improve grasp stability and form a closed loop controller, we add tactile sensors to the grippers.


\subsection{Measurement of parameters}
\begin{figure*}[t] 
  \begin{minipage}[]{0.4\textwidth} 
    \centering 
    \includegraphics[width=0.80\textwidth]{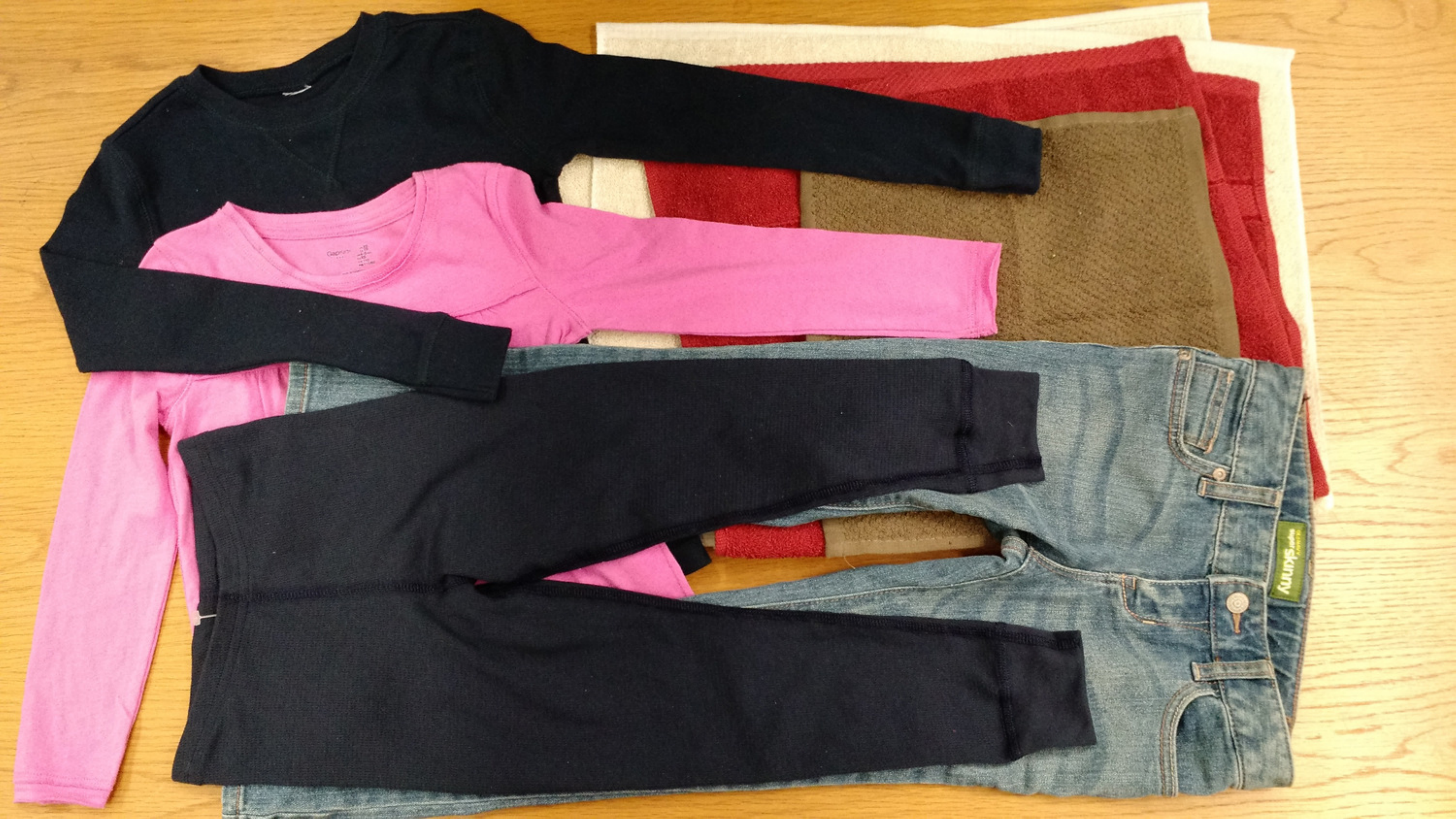} 
  \end{minipage}
  \begin{minipage}[]{0.6\textwidth} 
	\small
	\centering
	\footnotesize
    \begin{tabular}{c|p{0.8cm}|p{1.3cm}|p{1.6cm} | p{1.1cm}} \hline
      Garment Type &  \centering Stretch (\%) & \centering Friction Angle ($^{\circ}$) & \centering Maya Shear Resistance &\centering Maya Friction  \tabularnewline \hline
      Long-Sleeve T-Shirt (large) & \centering $2.9$  & \centering $24.3$  & \centering $200$ & \centering $0.7$ \tabularnewline
			Long-Sleeve T-Shirt (small) &  \centering  $2.9$ & \centering $24.7$ & \centering $200$ & \centering $0.7$ \tabularnewline
      Jeans &  \centering $2.9$  & \centering  $19.1$ & \centering $200$ & \centering $0.5$ \tabularnewline
			Pants &   \centering  $1.7$ & \centering  $21.9$ & \centering $340$ & \centering $0.6$ \tabularnewline
			Large Towel &  \centering $2.2$  & \centering  $18.7$ & \centering $260$ & \centering $0.5$ \tabularnewline
			Medium Towel &  \centering $3.1$  & \centering  $22.3$ & \centering $190$ & \centering $0.6$ \tabularnewline
			Small Towel &  \centering $1.1$  & \centering $24.3$  & \centering $530$ & \centering $0.7$ \tabularnewline \hline
			{\bf{Average}} & \centering ${\bf{2.4}}$ & \centering ${\bf{22.2}}$ & \centering $\bf{274}$ & \centering $\bf{0.6}$ \tabularnewline
      \hline
    \end{tabular}
  \end{minipage} 
	\caption{{\sc{Left:} }A picture of our test garments.
	{\sc{Right:}} Results for each unfolding test on the garments. 
	We show the results of stretch percentage, Friction angle of the table, and the corresponding parameters in Maya by each test.
	The last row shows the average of each measurement component.
	}
\label{fig:garment_stat}
\end{figure*}

To make the off-line simulation better approximate the real scenario, as described in Section~\ref{section:Simulation}, we manually measure the stretch resistance of each garment and friction on the table.
Figure~\ref{fig:garment_stat}, left shows a picture of all the test garments we used in different colors, sizes, and material properties.
Figure~\ref{fig:garment_stat}, right table shows the measured parameters of each test garment, including stretch percentage and Friction angle, and corresponding \emph{Maya} parameters.
For common garments, these parameters do not have a significant variance.
Therefore, we suggest that if researchers use simulators such as \emph{Maya}, the average values of each column are a reasonably good initialization.

\subsection{Garment manipulation and folding}

\begin{figure*}[t]
\begin{center}
 \includegraphics[width=0.99\textwidth]{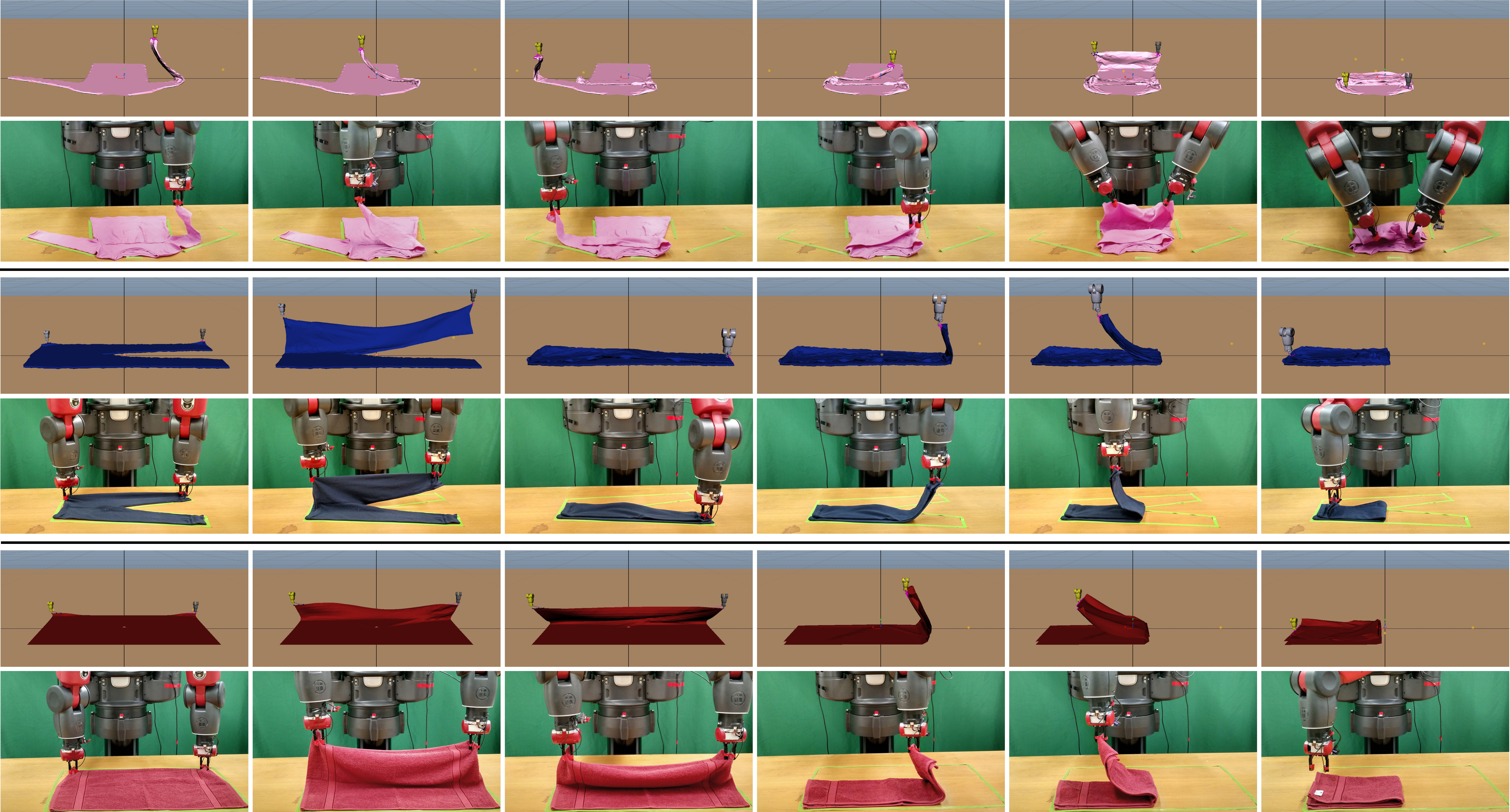}
\end{center}
   \caption{Successful folding examples with optimized folding trajectories from off-line simulation. 
	The first row of each group is from the simulation and the second row is from the real world (Green tape shows the original garment contour position).
	{\sc{Top Group:}} Long-sleeve shirt folding with $3$ steps.
	{\sc{Middle Group:}} Long pants folding with $2$ steps.
	{\sc{Bottom Group:}} Medium size towel folding with $2$ steps.}
\label{fig:good_examples}
\end{figure*}

Figure~\ref{fig:good_examples} shows three successful folding examples from the simulation and the real world, including a long-sleeve shirt, a pair of pants, and a medium size towel. 
We show six key frames for each folding task.
The folding poses from the simulation are in the first row of each group with an optimized trajectory.
We also show corresponding results from the real world.
The green tape contour on the table indicates the original position of the garment.


Each garment is first segmented from the background and key points are detected from the binary mask, which takes $3-5$ secs on a regular CPU.
The algorithm discussed in Sec.~\ref{sec:localization}.
Given the key points, a corresponding multi-step folding plan is created 
For each garment, we have optimized trajectories for each folding step.
Here, we map these optimized trajectories to our scenario according to the generated folding plan.
Then the Baxter robot follows the folding plan with optimized trajectories to fold the garment.
We can see that the deformation of the real garment and the simulated garment is very similar.
Therefore, the final folding outcome is comparable to the simulation.

Table~\ref{fig:garment_experiments_stat} shows statistical results of the garment folding test.
Each time one or two robotic arms fold the garment counts as one fold.
We run $10$ trials for each test garment.
It turns out that the folding performance of the Long-Sleeve T-Shirts and Towels are very stable with our optimized trajectories.
Jeans and pants are less stable because the shear resistance of the surface is relatively high, and sometimes is difficult to bend, leading to unsuccessful folding.
In the successful folding cases for jeans and pants, we sometimes ended up with small wrinkles, but the folding plan was still able to complete successfully.
We also show the average time to fold a garment in the last row. 
The robot is able to fold most garments in about $1.5$ minutes.

There is a trade-off between doing contour fitting at each step and total time spent to fold a garment.
In this work, we start with one template and then assume that each step after that the folded garment is close to that in the simulation.
Our results in Table~\ref{fig:garment_experiments_stat} verify that this method works well and is able to save time since we only do the contour fitting once.
With our simulated trajectories, the Baxter robot is able to fold a garment under predefined steps correctly.
An alternative method could use the contour fitting at each step but this would require more time and computation.

We note that some failures due to the motor control error from the Baxter robot.
When the robot executes an optimized trajectory, its arm suffers from a sudden drop or jitter.
Such actions will raise pull forces to the garment, leading to drift and inaccurate folding.
This can be solved by using an industrial level robotic arm with more accurate control.
We also note that failures can be recognized with the correct sensing suite, and we are currently investigating ways to effect online error recovery for such failures.
One difference between the simulation and the real world we found is that moving a point on the mesh in the simulation is different from using a gripper to grasp a small area of a real garment and move it.
In the future, we hope to be able to simulate a similar grasp effect for the trajectory optimization.

\begin{table}[t] 
	\centering
		\footnotesize
    \begin{tabular}{c|p{1.0cm}|p{1.6cm}|p{1.6cm} } \hline
      Garment Type &  \centering \# of folds & \centering Success Rate & \centering Avg. Time (sec) \tabularnewline \hline
      L-S T-Shirt (large) & \centering $3$  & \centering  $10/10$ & \centering  $121$  \tabularnewline
			L-S T-Shirt (small) &  \centering  $3$ & \centering $10/10$ & \centering  $118$  \tabularnewline
      Jeans &  \centering $2$  & \centering $7/10$  & \centering  $88$  \tabularnewline
			Pants &   \centering  $2$ & \centering  $8/10$ & \centering $88$ \tabularnewline
			Large Towel &  \centering $2$  & \centering  $10/10$ & \centering  $90$   \tabularnewline
			Medium Towel &  \centering $2$  & \centering  $10/10$ & \centering  $88$  \tabularnewline
			Small Towel &  \centering  $2$ & \centering  $10/10$ & \centering  $83$  \tabularnewline \hline
			{\bf{Average}} & \centering ${\bf{2.3}}$  & \centering ${\bf{9.3/10}}$ & \centering ${\bf{97}}$  \tabularnewline
      \hline
    \end{tabular}
	\caption{Results of folding test for each garment . We show the number of folding steps, successful rate, and total time of each garment. Each garment has been tested $10$ times. L-S stands for Long-Sleeve. The time is the average over all successful trials for each garment.
	}
\label{fig:garment_experiments_stat}
\end{table}

\section{Conclusion}

In this paper, we propose a novel solution to find an optimal trajectory for manipulation of deformable garments.
We first create a simulation environment that is comparable to the real world.
By minimizing a quadratic objective function that measures dissimilarity between simulated folded shape and user specified shape iteratively, we obtain an optimized trajectory.
The trajectory is then mapped to a real robot and executed accordingly.
Experimental results demonstrate that with our optimized trajectories, the Baxter robot can manipulate the garment efficiently and accurately.

\textbf{Acknowledgments} We'd like to thank J. Weisz and J. Varley, for many discussions. 
We'd also like to thank NVidia Corporation, Intel Corporation, and Takktile LLC for the hardware support. 
This material is based upon work supported by the National Science Foundation under Grant No. 1217904 and in part by the JSPS Postdoctoral Fellowships for Research Abroad.


\bibliographystyle{plain}

\end{document}